%% file: main.tex
\documentclass[10pt,twocolumn,letterpaper]{article}

\usepackage[pagenumbers]{cvpr} 

\input{preamble}

\definecolor{cvprblue}{rgb}{0.21,0.49,0.74}
\usepackage[pagebackref,breaklinks,colorlinks,citecolor=cvprblue]{hyperref}

\newcommand*{\affaddr}[1]{#1} 
\newcommand*{\affmark}[1][*]{\textsuperscript{#1}}
\newcommand*{\email}[1]{\texttt{#1}}

\def\paperID{*****} 
\def\confName{CVPR}
\def\confYear{2024}

\title{Q-Seg: Quantum Annealing-Based Unsupervised Image Segmentation}

\author{%
Supreeth Mysore Venkatesh\affmark[1,3], Antonio Macaluso\affmark[1], Marlon Nuske\affmark[2], Matthias Klusch\affmark[1], Andreas Dengel\affmark[2,3]\\
\affaddr{\affmark[1]\textit{German Research Center for Artificial Intelligence (DFKI), Saarbruecken, Germany}}\\
\affaddr{\affmark[2]\textit{German Research Center for Artificial Intelligence (DFKI), Kaiserslautern, Germany}}\\
\affaddr{\affmark[3]\textit{University of Kaiserslautern-Landau (RPTU), Kaiserslautern, Germany}}\\
\email{\{supreeth.mysore, antonio.macaluso, marlon.nuske,}\\
\email{matthias.klusch, andreas.dengel\}@dfki.de}\\
}

\begin{document}
\maketitle
\input{sec/0_abstract}

\input{sec/all_sections}
{
    \small
    \bibliographystyle{ieeenat_fullname}
    \bibliography{main}
}

\input{sec/X_suppl}

\end{document}

%% file: preamble.tex
\usepackage[dvipsnames]{xcolor}

\usepackage{graphicx}
\usepackage{amsmath}
\usepackage{amssymb}
\usepackage{booktabs} 
\usepackage{algorithm}
\usepackage{algpseudocode}
\usepackage{xr}
\usepackage{tabularx}

\usepackage{booktabs} 
\usepackage{colortbl} 

\definecolor{headercolor}{rgb}{0.65, 0.65, 0.65}
\definecolor{rowcolor}{rgb}{0.9, 0.9, 0.9}

\usepackage{microtype}

\usepackage{algpseudocode}
\usepackage{algorithm}

\newcommand{\newtext}[1]{{\color{black}{#1}}}

%% file: sec/0_abstract.tex
\begin{abstract}
We present Q-Seg, a novel unsupervised image segmentation method based on quantum annealing, tailored for existing quantum hardware. We formulate the pixel-wise segmentation problem, which assimilates spectral and spatial information of the image, as a graph-cut optimization task. Our method efficiently leverages the interconnected qubit topology of the D-Wave Advantage device, offering superior scalability over existing quantum approaches and outperforming several tested state-of-the-art classical methods. Empirical evaluations on synthetic datasets have shown that Q-Seg has better runtime performance than the state-of-the-art classical optimizer Gurobi. The method has also been tested on earth observation image segmentation, a critical area with noisy and unreliable annotations. In the era of noisy intermediate-scale quantum, Q-Seg emerges as a reliable contender for real-world applications in comparison to advanced techniques like Segment Anything. Consequently, Q-Seg offers a promising solution using available quantum hardware, especially in situations constrained by limited labeled data and the need for efficient computational runtime.
\end{abstract}

%% file: sec/all_sections.tex
\section{Introduction}

Image segmentation is pivotal in computer vision, enabling pattern recognition across diverse applications including medical imaging, autonomous navigation, search engines, and earth observation.
Technically, segmenting an image consists of partitioning it into non-overlapping regions with distinct attributes, aligning with human perception.
Algorithmic advancements in computer vision research, leading to substantial performance enhancements, are closely linked to underlying hardware capabilities.

\begin{figure}[ht]
\centering
\includegraphics[width=0.47\textwidth]{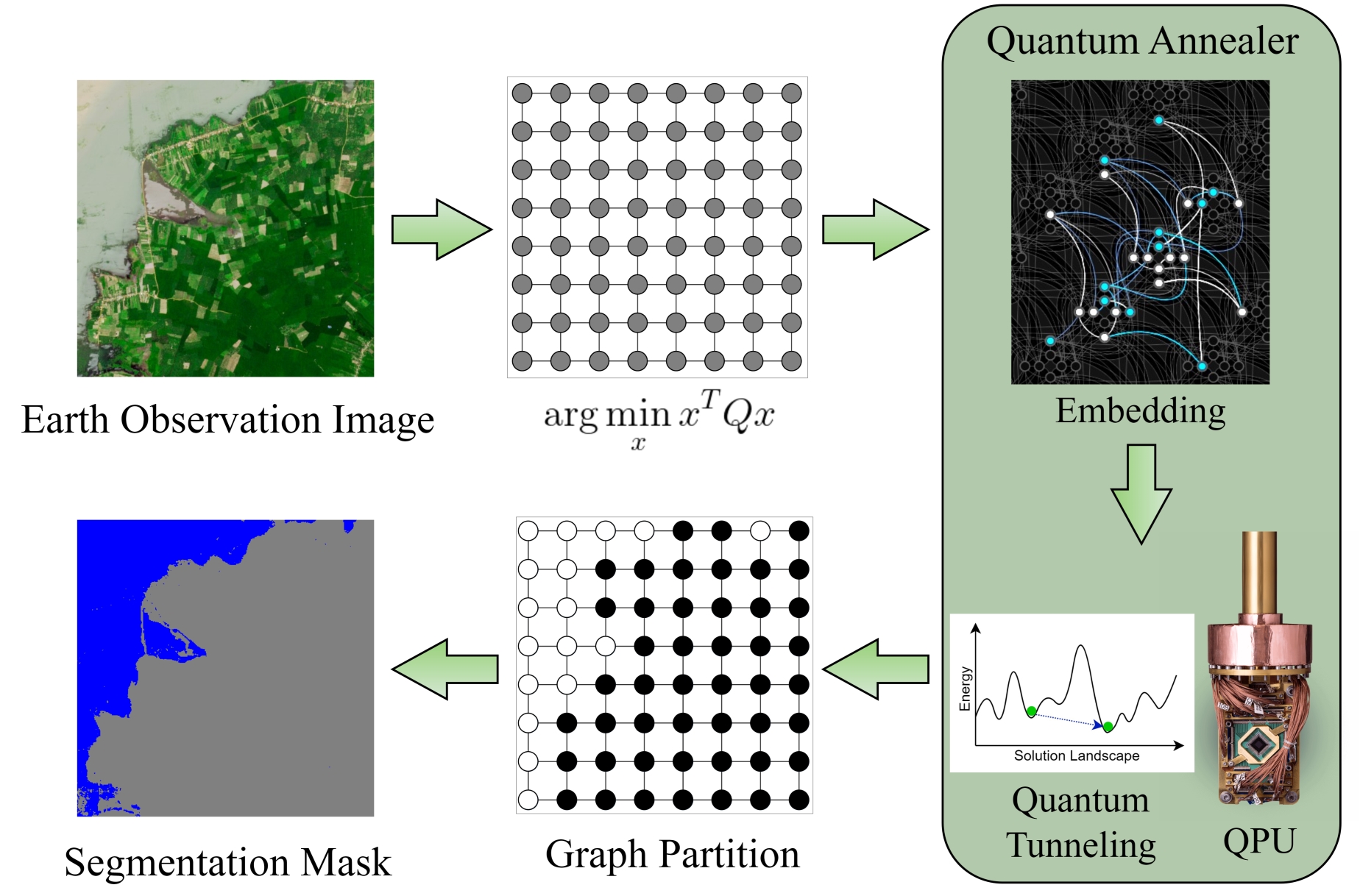}
\caption{The proposed \textit{Q-Seg} for unsupervised image segmentation generates a graph representing the original image. The distinct semantic regions in the image are identified by finding the minimum cut in the graph, an NP-hard problem that can be formulated as Quadratic Unconstrained Binary Optimization (QUBO). The QUBO problem is subsequently encoded into the physical topology of the D-Wave quantum annealer. Leveraging quantum tunneling, the annealing process efficiently explores an exponentially large solution space to locate the global optimum, which can be decoded as the segmentation mask.}
\vspace*{-10pt}
\label{fig: poster}
\end{figure}

Recently, quantum computing has emerged as a powerful alternative to solve complex real-world applications.
However, current quantum computers have limited capabilities, impacting algorithm design and testing. Specifically for computer vision, several quantum methods have been proposed for image classification, multi-object tracking, image matching, motion segmentation, edge detection and image segmentation.
Nonetheless, all these approaches do not show a concrete advantage for practical applications using existing quantum hardware.

In this paper, we introduce \textit{Q-Seg}, a novel unsupervised method that 
formulates the image segmentation problem as a graph-based optimization task, solvable via quantum annealing.
The vast solution space, exponential in pixel count, is efficiently navigated to yield near-optimal segmentation congruent with contemporary annealer architectures.
To show the ready-to-use capability of our approach, we run experiments using the \textit{D-Wave Advantage} annealer on synthetic and real-world satellite imagery datasets, specifically for Forest Cover and Flood Mapping.
In the case of Forest Cover segmentation, we obtain high-quality masks, underscoring the utility of unsupervised methods in scenarios with noisy or uncertain ground truth.
For Flood Mapping, we achieve near-optimal segmentation quality, rivaling supervised methods.
These results accentuate the capability of \textit{Q-Seg} to segment small and medium-sized images with existing quantum annealers, benchmarked against the classical state-of-the-art optimizer \textit{Gurobi} for runtime and quality assessment and also compared against \textit{Segment Anything Model (SAM)} on real-world images.

\section{Related Works} \label{sec: related works}

State-of-the-art methods for image segmentation primarily involve the use of Deep Convolutional Neural Networks in a supervised manner. These approaches can achieve remarkable segmentation performance when supplied with accurate annotations \cite{RFB15a}. However, in various domains acquiring labeled data encounters challenges such as the considerable cost and time required for manual annotation, the necessity for domain expertise, and scalability issues when handling large datasets. Also, subjectivity among annotators and variations in interpretation can introduce inconsistencies.

These limitations underscore the importance of unsupervised techniques, especially when leveraging vast unlabeled datasets \cite{Wang_2023}.
Unsupervised segmentation algorithms can be classified into several categories.
Thresholding is a pixel-wise segmentation method that assigns pixels to regions based on intensity values.
While computationally efficient, it is susceptible to issues arising from illumination variations and noise, which reduce its potency in situations where factors beyond intensity affect performance \cite{8933594}.
On the other hand, feature space clustering segments the spectral feature space into distinct clusters, each representing a region.
However, a notable limitation is its disregard for spatial relationships between pixels \cite{609410}.
Edge-based segmentation techniques detect discontinuities in image attributes, primarily pixel intensities, to delineate object boundaries.
While adept at handling images with clear object edges, this method struggles with noisy or indistinct boundaries, resulting in segmentation errors.
Region-based methods focus on pixel homogeneity. Techniques like region-growing initialize with seed pixels and expand regions based on similarity metrics until a set condition is satisfied.
Conversely, region-splitting dissects larger segments into finer clusters when they lack homogeneity in attributes like color, texture, or intensity \cite{kotaridis2021remote}.
However, region-based methods falter in scenarios where objects exhibit similar attributes or when similarity thresholds are not optimally set, resulting in over or under-segmentation \cite{NIPS2009_a7aeed74}.

Graph-based segmentation conceptualizes image segmentation as a graph partitioning task where nodes correspond to pixels or regions, while edges, weighted by pixel dissimilarity, link adjacent nodes \cite{yi2012image}.
The primary goal is to discern subgraphs that align with coherent clusters in the image.
A salient feature of this approach is its capacity to amalgamate both boundary and regional information, yielding globally optimal solutions for specific cases.
This makes it a potent tool for image segmentation, especially when prior data is sparse.
However, the inherent complexity of optimization tasks, such as graph cuts or clustering, renders the method computationally demanding \cite{8694009}.


Research has advanced beyond classical computing paradigms, predominantly exploring quantum algorithms for computer vision tasks \cite{9932103}.
Graph-based segmentation using variational quantum circuits has been investigated \cite{tse2018graph}.
However, the method has scalability issues in practical applications, as the qubit requirements, corresponding to the number of image pixels, limit their feasibility for large-scale use.

\newtext{A recent study \cite{presles2023synthetic} explores the use of quantum annealing for image segmentation by incorporating the Expectation Maximization (EM) algorithm into graph-based segmentation as a constrained optimization problem.}
This approach, while innovative, does not ensure that solutions to the reformulated unconstrained problem align with those of the initial constrained problem \cite{10.1007/978-3-031-04148-8_11}, and faces scalability issues due to the graph structure. 
Furthermore, the comparison of quantum annealer performance to classical methods lacks engagement with advanced segmentation techniques.
Conversely, \textit{Q-Seg} proposes a simplified graph structure, starting directly with an unconstrained optimization problem. It eliminates redundant solutions, offering a completely unsupervised approach, enhancing scalability and efficiency. 
Our experiments detail a thorough analysis of the physical qubit requirements, affirming \textit{Q-Seg}'s advancements.

\section{Preliminaries} \label{sec: preliminaries}

In contemporary quantum computing, two distinct paradigms have broadly emerged for tackling diverse problem sets: universal gate quantum computing and adiabatic quantum computing (AQC). 
This paper focuses on the adiabatic quantum computing paradigm, a computational approach that employs the quantum annealing process to address quadratic unconstrained binary optimization (QUBO) problems, defined as follows \cite{Glover2022}:

\begin{equation} \label{eqn: QUBO}
\arg \min_x \ x^T Q x = \arg \min_x  \sum_{i=1}^{n} l_{i}x_i + \sum_{1 \leq i < j \leq n} q_{ij} x_i x_j
\end{equation}
\vspace*{-1pt}
where $x \in \{0,1\}^n$ is the binary variable vector and $Q \in \mathbf{R}^{n \times n}$ is the QUBO matrix, with its diagonal and off-diagonal elements representing linear ($l_i$) and quadratic ($q_{ij}$) coefficients, respectively.
Despite its straightforward formulation, the QUBO problem is NP-hard, presenting significant computational challenges for large-scale instances since the solution space grows exponentially with the input size \cite{Glover2022}.
Moreover, unlike continuous optimization, the discrete optimization landscape here demonstrates non-linear cost deviations, i.e., small changes in input can lead to large variations in the cost function.

Quantum annealers, utilizing the principles of quantum mechanics, particularly the adiabatic theorem, efficiently navigate complex solution landscapes. 
Through quantum tunneling, these devices can potentially escape local minima, offering a promising alternative to classical optimization methods. 
However, the practical implementation of large QUBO problems on quantum annealers is hindered by limitations such as a finite number of qubits, connectivity constraints, susceptibility to noise, and sensitivity to optimization landscape and algorithmic characteristics. 
Thus, identifying suitable problems is essential for unlocking the full potential of quantum annealing in tackling complex optimization tasks.

\section{Methods} \label{sec: methods}

This section describes \textit{Q-Seg} in detail, delineating the algorithmic workflow, encapsulating the preprocessing, problem formulation, quantum annealing process, and how to obtain the segmentation mask.

\paragraph{Image to Graph:}
In graph-based segmentation, a graph is constructed with a node for each image pixel, typically supplemented by two terminal nodes termed \textit{source} and \textit{sink}, which serve as theoretical constructs for the object and the background. Edges linking adjacent nodes incorporate weights that reflect either similarity or dissimilarity. Additionally, the source and sink nodes connect to all pixel-representing nodes, with edge weights encoding prior knowledge. Evolving from conventional techniques, our strategy exploits the quantum annealer hardware's potential, presenting a formulation that achieves greater efficiency relative to classical counterparts. \textit{Q-Seg} begins by forming a lattice graph from the input image, adopting a grid-graph structure and omitting the terminal nodes. In this representation, graph nodes equate to image pixels, preserving the spatial inter-pixel connectivity crucial for segmentation. The choice of defining the edge weights is application-dependent, and different usecases may require distinct metrics tailored to specific characteristics of the image data. For scenarios with limited annotated samples, establishing optimal edge weights is approached as an optimization problem, detailed further in the appendix.

\newtext{In our approach, edge weights represent the similarity between neighboring pixels' intensity values, calculated using a Gaussian similarity metric. For neighboring pixels \( p_i \) and \( p_j \) with intensity values \( I(p_i) \) and \( I(p_j) \), respectively, we use the following function to compute the edge weight \cite{NIPS2001_801272ee}:

\begin{equation} \label{eqn: edge weight gaussian}
w'(p_i, p_j) = 1 - \exp\left(-\frac{(I(p_i) - I(p_j))^2}{2\sigma^2}\right)
\end{equation}
where \( \sigma \) is the standard deviation parameter that controls the spread of the Gaussian function. This similarity metric yields values in the range \([0, 1]\).

To ensure the edge weights are both positive and negative, we normalize these weights to the range $[-1, 1]$. This is done by first finding the minimum and maximum weights across all edges, and then applying the following normalization formula:
\begin{equation} \label{eqn: gaussian normalized}
    w(p_i, p_j) = -1 \times \left( \frac{(b-a) \cdot (w'(p_i, p_j) - \min(w))}{\max(w) - \min(w)} + a \right)
\end{equation}
where \( a = -1 \) and \( b = 1 \) are the desired range bounds. Normalizing the edge weights to include both positive and negative values is crucial as it ensures that the segmentation algorithm can effectively differentiate between similar and dissimilar pixel pairs.
This formulation of the edge weights provides a general similarity score that effectively works for the experiments in this paper, but every specific use case might require a custom edge weight metric.}

When representing an image as a graph, segmentation involves identifying a graph cut that partitions the vertices based on the similarity metric of edge weights. Specifically, when the edge weights represent pixel similarity, dividing the graph into semantically distinct regions can be framed as discovering the minimum cut in the graph, minimizing the sum of cut edges' overall similarity. Conversely, when edge weights measure pixel dissimilarity, the segmentation task targets the maximum cut. For real-valued edge weights, the maximum cut problem can be transformed into a minimum cut problem by negating the edge weights and vice-versa. Consider an undirected weighted graph \( G(V, w) \) with no self-loop edges, where \( V \) is the set of vertices and \( w: V \times V \to \mathbf{R} \) is the edge weight function. In this context, a 'cut' denotes the partitioning of the graph into disconnected subgraphs, which arises from the removal of a specific set of edges. The \textit{cost} of a cut is quantified as the aggregate of the weights of the edges removed as follows:

\vspace*{-12pt}
\begin{equation}
    \mathcal{MINCUT}(G) = \arg \min_{A,\overline{A}} \sum_{i \in A, j \in \overline{A}} w(v_{i},v_{j})
\end{equation}
\vspace*{-5pt}
$$\text{such that } A \cup \overline{A} = V, A \cap \overline{A} = \varnothing$$

\begin{figure}[ht]
\centering
\includegraphics[width=0.47\textwidth]{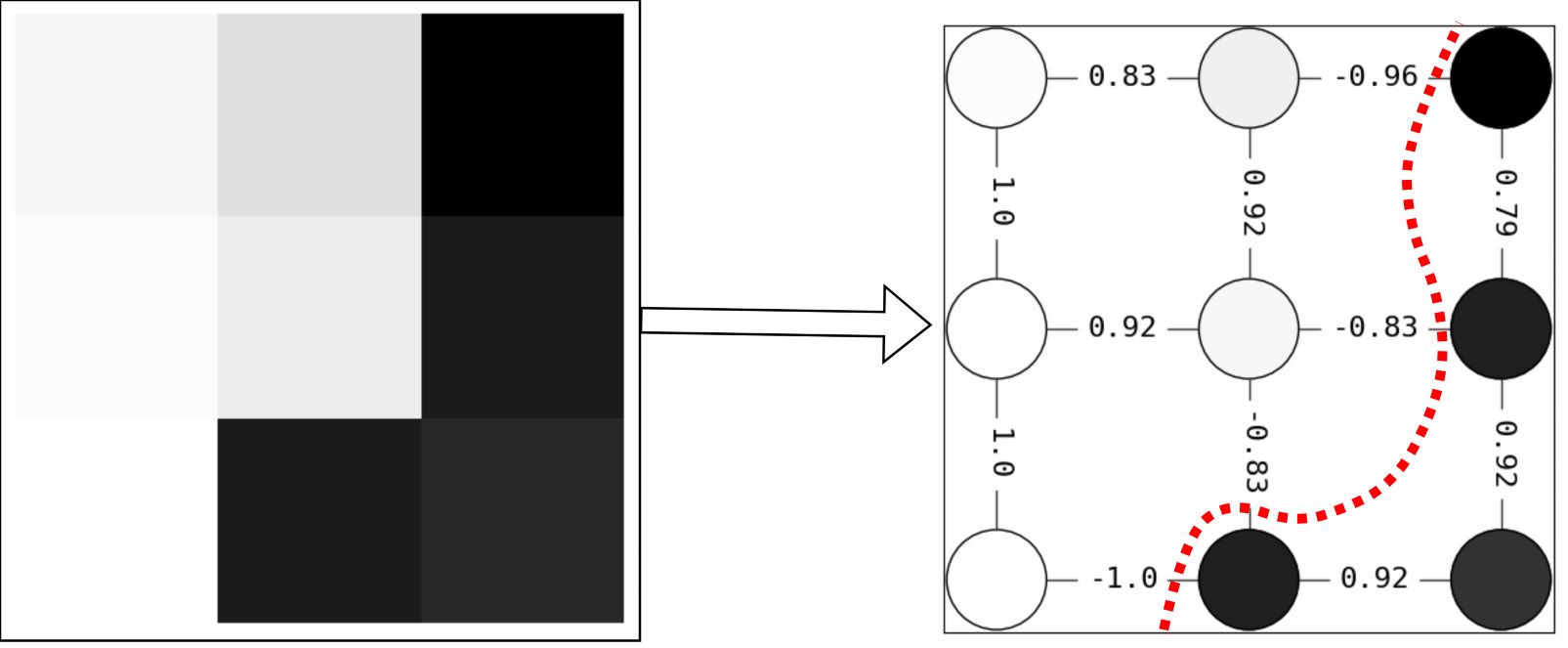}
\caption{Converting the pixel values into edge weights in the grid graph. The grid structure of the graph captures the spatial information (high-level information such as regions or objects) and the edge weight captures the spectral information (low-level information such as pixel values). The red-colored dotted curve passes through the set of edges that divide the vertices into two distinct sets, such that the sum of the edge weights is minimum.}%
\label{fig: image to gridgraph}
\end{figure}

\hspace*{-16pt}
where $w(v_{i},v_{j})$ is the weight of the edge connecting the nodes $v_i$ and $v_j$. In its general formulation where edge weights can be positive and negative, finding the minimum (or maximum) cut in the graph is NP-Hard. However, there are heuristic algorithms for finding minimum cuts in planar graphs in polynomial time for specific instances \cite{doi:10.1137/0204019}. Nevertheless, for the generic case, the problem still remains NP-complete \cite{MONIEN1988209}. Indeed, for an image with $n$ pixels, there exist $2^n$ potential solutions, rendering exhaustive enumeration of all possible solutions infeasible. Fig. \ref{fig: image to gridgraph} illustrates converting a grey image of size $3 \times 3$ to a grid graph where the edge weights $\in [-1,1]$ represent the similarity between the neighboring pixels, as described by Eq. \ref{eqn: gaussian normalized}.

\paragraph{Formulating QUBO:}
To overcome the computational hurdles on a classical computer, and be able to segment an image efficiently, we use a quantum annealer.
This necessitates recasting the minimum cut problem into QUBO form, which is compatible with the operational framework of quantum annealing.
A cut is also conceptualized as separating the vertex set into two exclusive subsets, with a minimum cut characterized by the least possible sum of edge weights bridging these subsets.
If $G$ has $n$ nodes, associating a binary variable $x_{v_{i}}$ to vertex $v_i$ for all $i \in \{1,2,...,n\}$ allows to encode the solution as $v_i \in A \ \forall \ x_{v_{i}}=0$  and $v_i \in \overline{A} \ \forall \ x_{v_{i}}=1$ or vise-versa.
The objective function for computing the minimum cut in this binary framework is articulated as \cite{Glover2022}:

\begin{equation}\label{eqn: qubo cost function}
    x^* = \arg \min_x \sum_{1 \leq i < j \leq n} x_{v_{i}} (1-x_{v_{j}}) w(v_{i},v_{j})
\end{equation}
\vspace*{-5pt}
$$\text{where } x = x_{v_{1}}, x_{v_{2}},...,x_{v_{n}} \text{and } x_{v_{i}} \in \{0,1\}$$

The cost function in Eq. \eqref{eqn: qubo cost function} is equivalent to the QUBO formulation in Eq. \eqref{eqn: QUBO} \cite{gcs-q}.
Given the edge weights $w(v_{i},v_{j})$, substituting them into Eq. \eqref{eqn: qubo cost function} and simplifying yields a quadratic expression in binary variables $x_{v_{i}}$ where $i \in \{1,2,...,n\}$.
The construction of the $Q$ matrix in Eq.\eqref{eqn: QUBO}, is derived from the coefficients of terms in the quadratic equation. 
Note that the QUBO formulation of the minimum cut does not introduce any overhead in terms of the number of binary variables compared to the input size in the original problem, as is the case with other existing formulations \cite{bilp-q, presles2023synthetic}.

\paragraph{QUBO with Quantum Annealing:}
Quantum annealing offers a probabilistic approach to solving the QUBO problem through the exploitation of quantum tunneling and entanglement.
The process begins by mapping the QUBO onto the topology of physical qubits, aligning the solution with the binary states of the qubits.
Throughout the annealing process, the system evolves from a superposition of all possible states toward the lowest energy state, corresponding to a state where the optimal QUBO solution has a high probability of being measured.
This evolution is governed by the annealing schedule, which carefully balances quantum properties and classical interactions to avoid local minima.

The \textit{D-Wave} quantum annealer, in particular, employs a specialized hardware design conducive to this process, making it well-suited for graph-based problems like image segmentation.
\newtext{As an annealer is a probabilistic machine, typically an execution involves several runs, and it samples a possible solution at each run.}
Through iterative adjustments of the system's parameters and repeated annealing cycles, the probability of discovering the global minimum increases.
The final readout produces a binary string directly translating into the segmented image.

To summarize, the \textit{Q-Seg} approach is described in Fig. \ref{fig: poster}, and the pseudo-code is shown in Algorithm \ref{algo:qgs}.


\begin{algorithm}[ht]
\caption{Outline of \textit{Q-Seg}}\label{algo:qgs}
\begin{algorithmic}[1] 
\Require Image $I$ with $n$ pixels
\Ensure Segmentation mask $M$ matching $I$'s dimensions
\State Construct grid graph $G(V,w)$ from image $I$
\State Formulate QUBO for minimum cut on $G$ (Eq. \eqref{eqn: qubo cost function})
\State Map the QUBO to the quantum annealer's architecture
\State Run the quantum annealing
\State Extract the lowest-energy sample $X^*$
\State Decode sample $X^*$ into segmentation mask $M$
\end{algorithmic}
\end{algorithm}

For an input image $I$, we construct a grid graph $G$ with a one-to-one correspondence between pixels in $I$ and vertices in $G$.
Edge weights $w(v_{i},v_{j})$ are assigned based on pixel similarity between $v_i$ and $v_j$, to segment $I$ by identifying a minimum cut in $G$.
The QUBO formulation captures this objective, translating the problem into a format amenable to quantum annealing.
The annealing process is iterated multiple times to ensure the reliability of the solution, which is then decoded into a binary mask $M$ as $M_{i} = x_{v_{i}}$ for each vertex $v_{i} \in V$, describing the regions in $I$.

The structure of the graph in our problem formulation, with nodes exhibiting a maximum degree of $4$, naturally fits the architecture of the existing quantum annealer hardware.
This congruence facilitates efficient mapping of logical to physical qubits, reducing the number of cloned qubits and inter-qubit coupling.

\section{Experiments} \label{sec: experiments}

In this section, we present experimental evaluations of \textit{Q-Seg} against state-of-the-art classical methods on a synthetic dataset and two practical usecases in the domain of earth observation.

\subsection{Experimental settings}

\subsubsection{Datasets}

\textbf{Synthetic Data:}
The complex operation in \textit{Q-Seg} involves computing the minimum cut in a grid graph, a key step essential for effective image segmentation.
We generate synthetic datasets comprising undirected, connected, and weighted grid graphs of square structure, with sizes ranging from $2 \times 2$ up to $44 \times 44$. This data allows assessing the performance of \textit{Q-Seg} in terms of runtime while increasing the problem size. 
For reproducibility, we employ a fixed seed to assign edge weights, which are randomly selected from a uniform distribution within the interval $[-1,1]$ ensuring that the edge weights are both, positive and negative.
Tests are conducted on five sets of graphs, each initialized with a different seed.

\textbf{Forest Cover:}
Forest segmentation is crucial for ecological sustainability enabling accurate assessment of deforestation rates, forest density, and changes due to climatic or human factors.
Due to the limited availability of time resources of the real quantum device, we randomly sample a subset of images from the \textit{DeepGlobe18} dataset\footnote{\href{https://www.kaggle.com/datasets/quadeer15sh/augmented-forest-segmentation}{https://www.kaggle.com/datasets/quadeer15sh/augmented-forest-segmentation}}, each measuring $256 \times 256$ pixels, accompanied by binary segmentation mask.
Furthermore, the original images are downscaled to $32 \times 32$ pixels, to fit the actual number of qubits in the real device. In particular, the preprocessing steps included median blurring and conversion to the HSV (Hue Saturation Value) color scale, tailored to the specific usecase.
For generating the input graph for \textit{Q-Seg} algorithm, we incorporate the Gaussian similarity \newtext{(Eq. \ref{eqn: edge weight gaussian})} as the edge weight metric, normalizing them within the range of $[-1, 1]$ \newtext{(Eq. \ref{eqn: gaussian normalized})}.

\textbf{Flood Mapping:}
The \textit{Sen1Floods11} dataset\footnote{\href{https://github.com/cloudtostreet/Sen1Floods11}{https://github.com/cloudtostreet/Sen1Floods11}}, designed for water segmentation in satellite imagery, comprises georeferenced samples with dimensions of $512 \times 512$ pixels across $13$ spectral bands.
The ground truth masks include binary labels and an additional value of $-1$, indicating uncertain areas, as annotated by experts.
Specifically, the subset of images used comprises pictures of the Bolivia region obtained from \cite{iselborn2023importance}.
The images are segmented at a pixel level, and then we work with $32 \times 32$ patches and subsequently integrate these patches to form the complete segmentation mask. This approach allowed for detailed analysis without compromising image integrity.

As preprocessing and customizing input for the subsequent task is typical for an unsupervised algorithm when applying it to a real-world usecase, likewise we follow a few steps for generating the graph for \textit{Q-Seg}.
We first calculated the normalized difference water index of the input images and \newtext{used a normalized Gaussian similarity metric as discussed in Eq. \ref{eqn: gaussian normalized}.}


\begin{figure*}[ht]
\centering
\includegraphics[width=\textwidth]{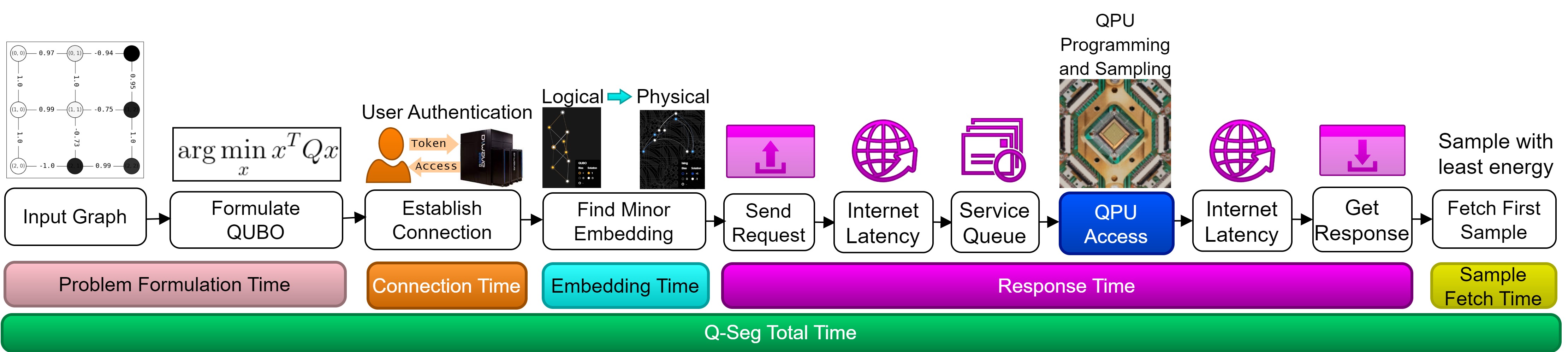}
\caption{The figure illustrates the operational pipeline of \textit{D-Wave} quantum annealer. It begins with reformulating the minimum cut problem as a QUBO, followed by authentication using a private token for remote access. The \textit{minorminer} tool maps the logical qubits in QUBO to the physical qubits in hardware. The problem instance is sent over the internet to join the queue on the shared \textit{D-Wave} device. The Quantum Processing Unit (QPU) performs the annealing process, producing a set of samples. The final step is extracting the optimal solution, identified by its lowest energy state that encodes the segmentation mask.}
\label{fig: annealer pipeline}
\end{figure*}

\subsubsection{Solving methods}
Our experiments were conducted using the \textit{D-Wave Advantage} quantum annealer for the graph-cut optimization, accessed through a cloud service via the $dimod$ \footnote{\href{https://docs.ocean.dwavesys.com/en/stable/docs_dimod/}{https://docs.ocean.dwavesys.com/en/stable/docs\_dimod/}} library. This device comprises $5670$ physical qubits and has \textit{pegasus} topology.
We evaluated \textit{Q-Seg} against the state-of-the-art solver \textit{Gurobi}.
For benchmarking against classical approaches, we implemented the \textit{Gurobi v\hspace{1pt}9.5.2} optimizer to solve the QUBO problem representing the minimum cut.
Furthermore, to quantitatively assess the segmentation quality, we compare the results of \textit{Q-Seg} for the Flood Mapping dataset with a state-of-the-art supervised method that uses gradient-boosted decision trees (\textit{GBDT}) \cite{iselborn2023importance}.
The classical part of our experiments was conducted on a system equipped with an \textit{Intel(R) Xeon(R)} CPU @ \textit{2.20GHz} and \textit{8 GB} RAM, with all software developed in \textit{Python 3.9}.

For real-world datasets, we also compare our results with the recent State-of-the-Art segmentation method \textit{SAM}, known for its robust zero-shot capabilities across various domains \cite{kirillov2023segany}.
To maintain a fair comparison with our unsupervised method, we employed the latest version of \textit{ViT-H}\footnote{\href{https://github.com/facebookresearch/segment-anything}{https://github.com/facebookresearch/segment-anything}.} model pretrained on \textit{SA-1B} \footnote{\href{https://ai.meta.com/datasets/segment-anything-downloads/}{https://ai.meta.com/datasets/segment-anything-downloads/}} dataset containing $11$ million images and $1.1$ billion masks, without specifically fine-tuning the model weights for the target usecase.
We allowed \textit{SAM} to generate $3$ segmentation masks for each image and the one with the highest confidence score is illustrated in the results.

In our experiments, we explored an alternative approach to solving the minimum cut problem with \textit{Gurobi}, adopting a constrained optimization mode. This approach, focusing on edge encoding, increases variable count significantly and resulted in poorer performance. 
Furthermore, we considered comparing our quantum annealing results with simulated annealing, often regarded as its direct competitor. However, research indicates simulated annealing falls short of quantum annealing in both speed and solution quality, often by exponential margins, leading us to omit it from our comparative analysis.

Finally, we omit the experimental comparison with the recently introduced segmentation method using quantum annealers \cite{presles2023synthetic} for three reasons.
First, the method works on pivotal assumptions for implementing the EM algorithm and requires prior knowledge about the distribution of pixel values of different regions, making it not fully unsupervised.
Second, the increased overhead in terms of the number of physical qubits hinders the possibility of running larger problem instances as those implemented to test \textit{Q-Seg}.
Third, the code is not made publicly available.
Nevertheless, we present a scalability analysis based on the number of physical qubits required by the existing quantum hardware and compare it with that of \textit{Q-Seg} (see Fig. \ref{fig: q-seg vs arxiv}).

\subsubsection{Metrics} \label{sec: metrics}

We evaluate our method using three different types of metrics: \textit{runtime}, \textit{quality} and \textit{scalability}. 

The \textit{runtime} allows us to estimate the efficiency of \textit{Q-Seg} in comparison to classical state-of-the-art optimizers. In Fig. \ref{fig: annealer pipeline}, a comprehensive description of all components of the quantum annealing pipeline is presented, emphasizing their significance when solving a problem on \textit{D-Wave} annealers that is accessible via the cloud.

In terms of \textit{quality}, we evaluate \textit{Q-Seg} using a set of established metrics. These include \textit{Intersection over Union (IoU)} to assess the accuracy of overlap in segmented regions, \textit{Accuracy} for determining the correct segmentation of pixels, \textit{Recall}, and \textit{Precision} to evaluate the detection and accuracy of segmentation.

Additionally, to assess the segmentation carried out by \textit{Q-Seg} in comparison to deterministic methods, we conducted a comparison of minimum cut values, employing the relative error as the metric for evaluation, defined as:
\begin{equation}
    E_r = \left| \frac{V_G-V_{QA}}{V_G} \right|
\end{equation}
where $V_G$ represents the minimum cut value from \textit{Gurobi} and $V_{QA}$ from the \textit{D-Wave Advantage} annealer.

Finally, we assess the \textit{scalability} of \textit{Q-Seg} by quantifying \textit{Qubit Complexity}, i.e., the number of physical qubits required in a real quantum device with respect to other existing quantum-annealing-based approaches.

\subsection{Results}
\subsubsection{Runtime Analysis}
\begin{figure}[ht]
\centering
\includegraphics[width=0.47\textwidth]{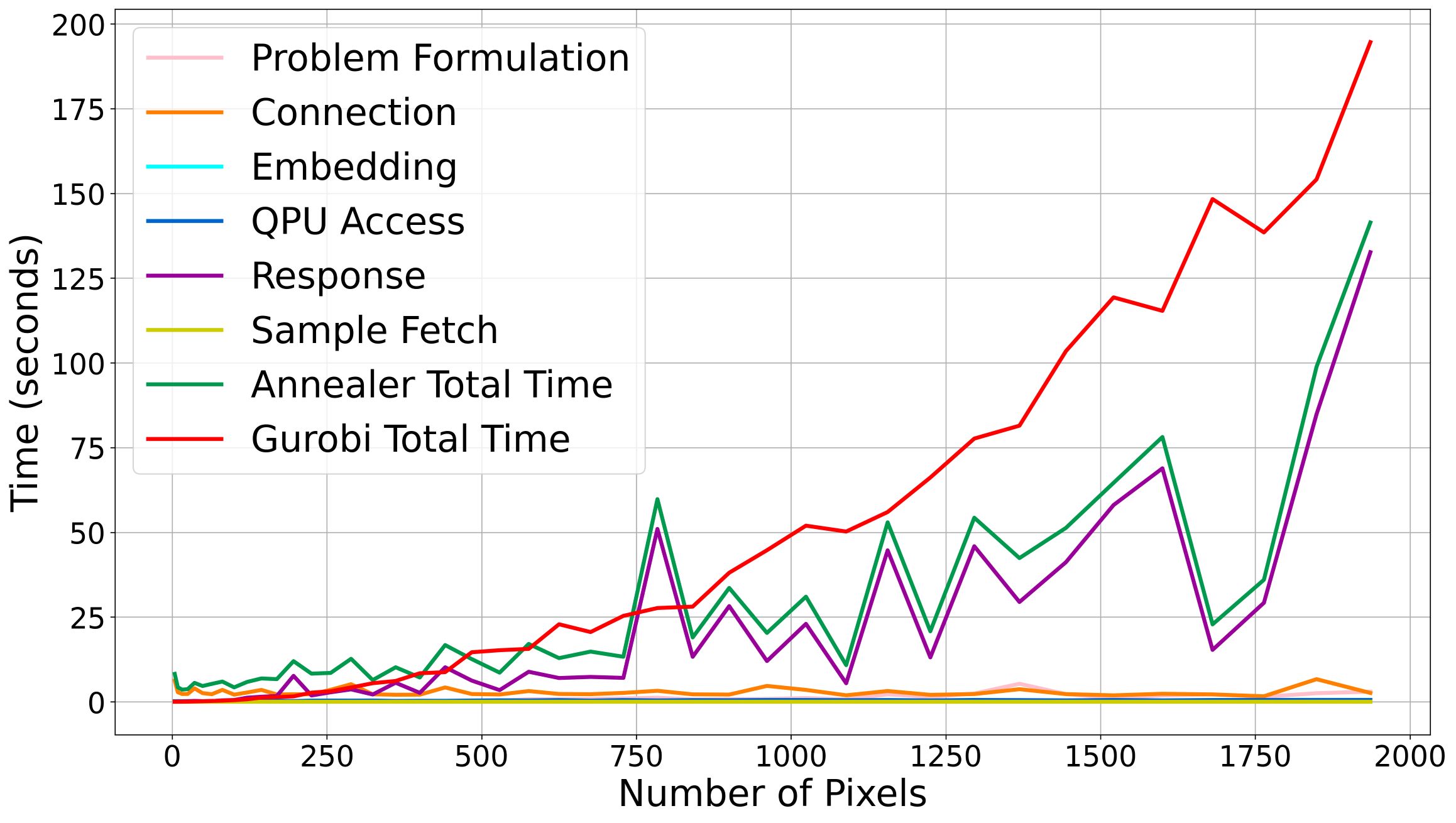}
\caption{Runtime comparison between \textit{D-Wave Advantage} annealer and \textit{Gurobi} for the synthetic data with $seed=333$. This graph illustrates the breakdown of annealer runtime, highlighting the components in Fig. \ref{fig: annealer pipeline}.
Despite shared and remote access, the \textit{D-Wave Advantage} demonstrates a consistently shorter total runtime compared to \textit{Gurobi}'s local execution.}
\label{fig: runtimes}
\end{figure}
We executed \textit{Q-Seg} on the synthetic dataset using both the \textit{D-Wave Advantage} and \textit{Gurobi} to compare classical and quantum algorithms while progressively increasing the problem size. The results, which encompass all the distinct components in the quantum annealing pipeline (Fig. \ref{fig: annealer pipeline}), are illustrated in Fig. \ref{fig: runtimes}. We observe that even when factoring in the runtime, which includes the response time from sending the problem to receiving the solution from the \textit{D-Wave} annealer—involving internet latency, queuing due to shared annealer usage, and the annealing process—the total time-to-solution for quantum annealing is significantly lower than that of \textit{Gurobi} that executes locally.

In a hypothetical scenario with a dedicated QPU, we could omit the connection establishment, internet latency, and queuing, which are significant contributors to the total runtime.
This assumption allows for a more direct comparison of the annealer's efficiency against \textit{Gurobi}.
In particular, \textit{QPU Access Time} reflects the Quantum Processing Unit's active problem-solving duration. This metric can be divided into \textit{QPU Programming Time}, \textit{QPU Sampling Time}, and \textit{QPU Access Overhead Time}.
The programming phase sets up the QUBO model in the annealer, sampling involves the actual annealing process, and the overhead time is associated with post-processing the samples.

\begin{figure}[ht]
\centering
\includegraphics[width=0.47\textwidth]{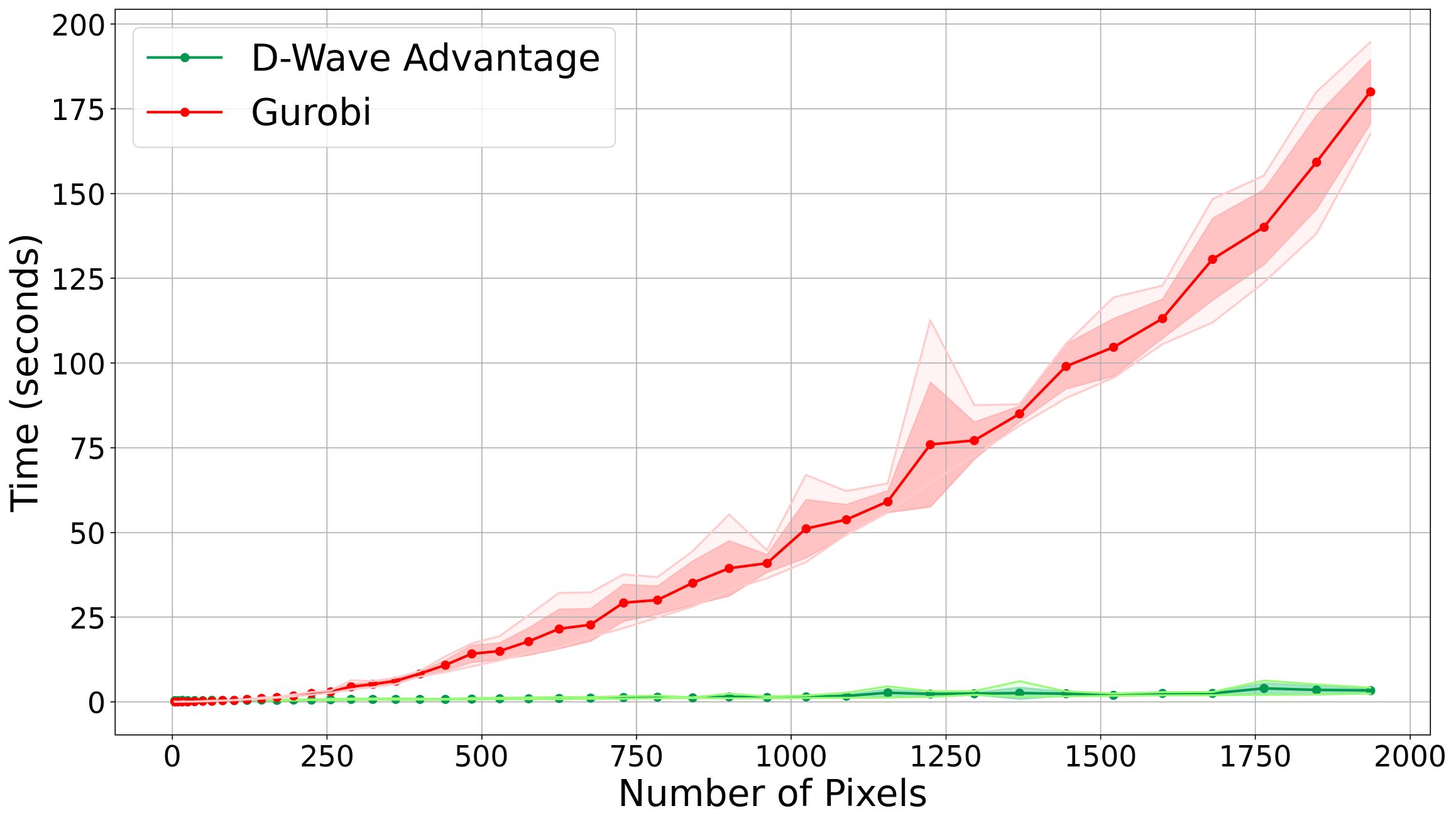}
\caption{The plot illustrates the mean runtime (represented by a solid line), the range of runtimes (indicated by the broadly shaded area), and the standard deviation (denoted by the lightly shaded area) is aggregated over five sets of synthetic data for the minimum cut operation on graphs of square images of varying sizes.
The results highlight the efficiency of \textit{Q-Seg}, especially evident in processing larger images.}
\label{fig: practical runtimes}
\end{figure}

Fig. \ref{fig: practical runtimes} compares the runtime of \textit{D-Wave Advantage} and \textit{Gurobi} assuming a dedicated onsite QPU. The hybrid process, involving QUBO formulation, embedding, annealing, and solution retrieval, exhibits even more pronounced superior performance compared to the previous scenario, especially as the problem size increases. This efficiency can be attributed to the grid graph structure aligning well with the QPU architecture. The newer \textit{D-Wave} devices, featuring enhanced qubit connectivity, further enhance this efficiency, making \textit{Q-Seg} particularly effective on these platforms.

\subsubsection{Quality}

\begin{figure}[ht]
\centering
\includegraphics[width=0.47\textwidth]{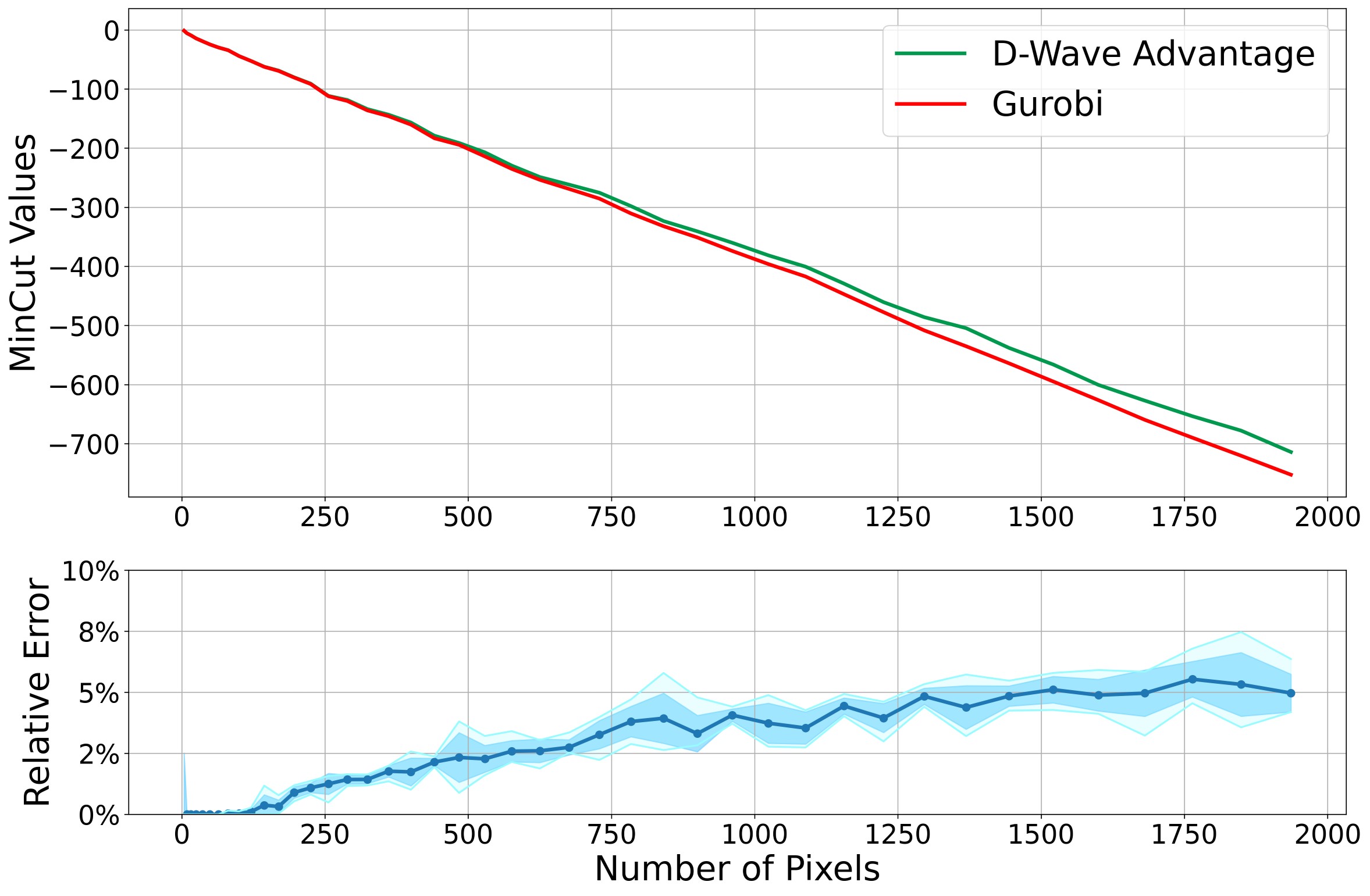}
\caption{The first plot contrasts the minimum cut values obtained by \textit{Gurobi} and the \textit{D-Wave Advantage} annealer on synthetic data with $seed = 333$, highlighting \textit{Gurobi}'s marginally better performance.
The trend of diminishing values substantiates the rising count of negative edges that are cut as the problem size increases.
To substantiate a consistent behavior, the second plot shows the mean (represented by the solid line), range (depicted by the broadly shaded area) and standard deviation (illustrated by the narrow shaded area) of the relative errors in the annealer solutions compared to \textit{Gurobi} aggregated over five sets of the synthetic data.}
\label{fig: annealer vs gurobi quality}
\end{figure}

While \textit{D-Wave} exhibits faster processing than \textit{Gurobi}, it is essential to evaluate the quality of the solutions. The analysis depicted in Fig. \ref{fig: annealer vs gurobi quality} shows that although \textit{Gurobi} consistently produces superior solutions, the margin is minimal. This observation is further supported by the relative error trends across different problem sizes.

Our approach demonstrates the ability to address large-scale problems with commendable quality using current quantum hardware. With an onsite annealer, the limitations on the number of annealing cycles, impacting the quality of solutions, could be mitigated, potentially further elevating the solution quality.

\paragraph{Forest Cover} \label{paragraph: forest cover}

\begin{figure}[ht]
\centering
\includegraphics[width=0.47\textwidth]{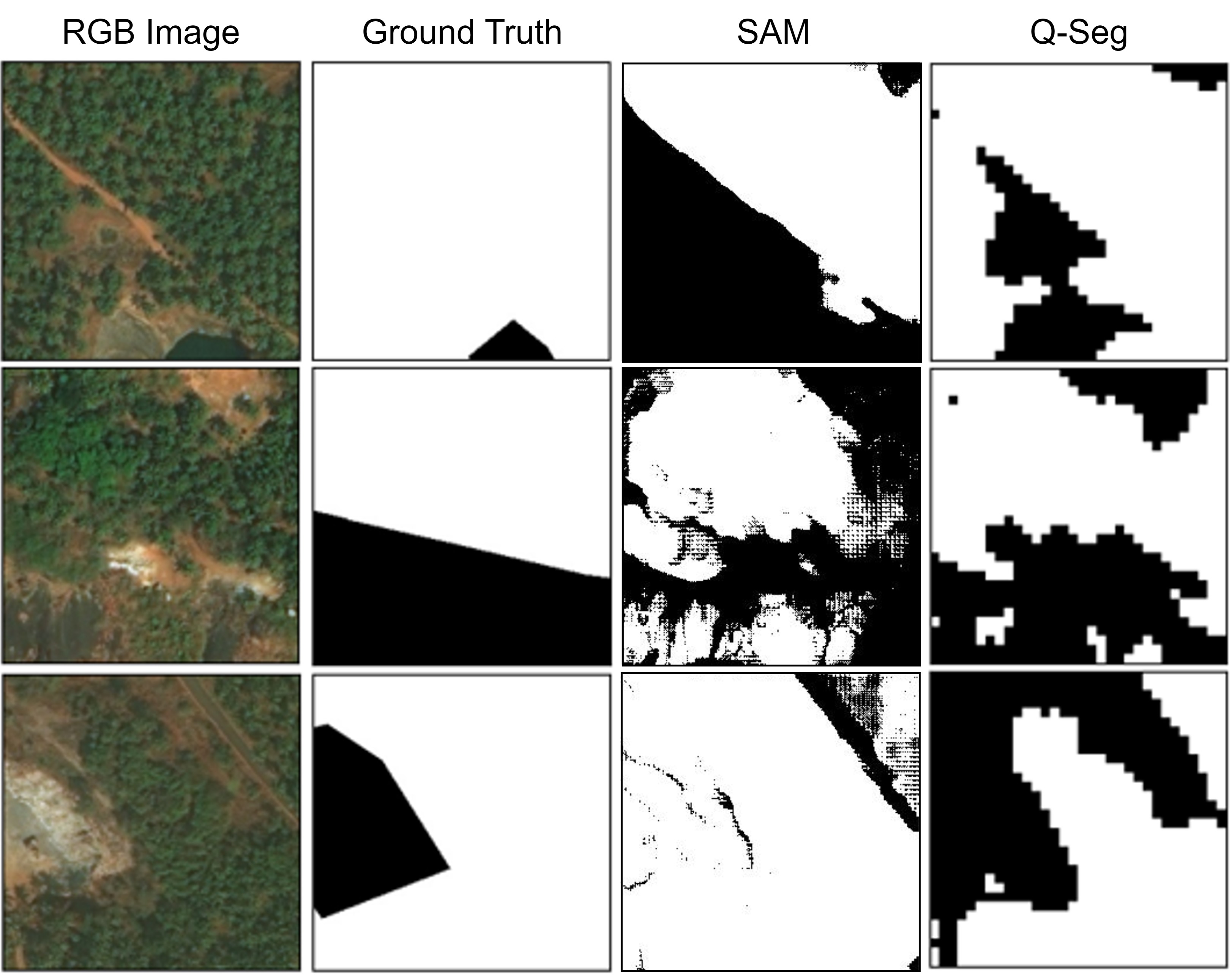}
\caption{Visualization of RGB images, ground truth masks, \textit{Q-Seg}, and \textit{SAM} segmentation results in the Forest Cover usecase. The ground truth masks show noisy labeling and irregularities, while the \textit{Q-Seg} results reveal more coherent and visually refined segmentation of Forest Cover, demonstrating notable improvements over \textit{SAM} in handling noisy labels.}
\label{fig: forest cover results}
\end{figure}

Segmenting Forest Cover poses a significant challenge, primarily attributed to the presence of noisy and inconsistent ground truth labels in the dataset.
This inconsistency can introduce bias during the training of supervised methods, potentially leading to misleading comparisons between algorithmic outcomes and ground truth masks.
In addressing these challenges, we compared the results of \textit{Q-Seg} with that of \textit{SAM}.
Fig. \ref{fig: forest cover results} provides a qualitative glimpse into the segmentation results of \textit{Q-Seg} against existing masks and contrasts these with the outcomes from the \textit{SAM} model.

Even state-of-the-art methods like \textit{SAM} fail to generalize for specific usecases, and even harder for datasets with noisy labels.
Our approach showcases its potential to provide more reliable and visually consistent segmentation results in such challenging environments.
Importantly, the consistency of poor-quality ground truth and the corresponding high-quality masks provided by \textit{Q-Seg} is congruous across most images in the dataset\footnote{\href{https://github.com/supreethmv/Q-Seg}{https://github.com/supreethmv/Q-Seg}}.
By preprocessing the image adapting to the usecase and using a suitable edge weight metric, \textit{Q-Seg} emerges as a better contender for unseen images when compared to existing state-of-the-art techniques, including \textit{SAM}.
Since the Forest Cover dataset is not well labeled, a quantitative comparison of the results of \textit{Q-Seg} with the noisy labels would not be a fair measure and thus we extend our experiments with the Flood Mapping usecase.
As a takeaway, the results demonstrate the effectiveness of \textit{Q-Seg} making graph-based techniques feasible by leveraging the capabilities of quantum annealers.

\paragraph{Flood Mapping}

Commonly, state-of-the-art methods for image segmentation rely on supervised learning, utilizing annotated images to train algorithms for correctly identifying masks of unlabelled images. In the context of the flood mapping dataset, the \textit{GBDT} stands out as the state-of-the-art method, even outperforming CNNs in this specific case \cite{iselborn2023importance}. Alongside \textit{GBDT}, we include a comparison with \textit{SAM}, using its pretrained model to assess performance in this usecase.

\begin{figure}[ht]
\centering
\includegraphics[width=0.47\textwidth]{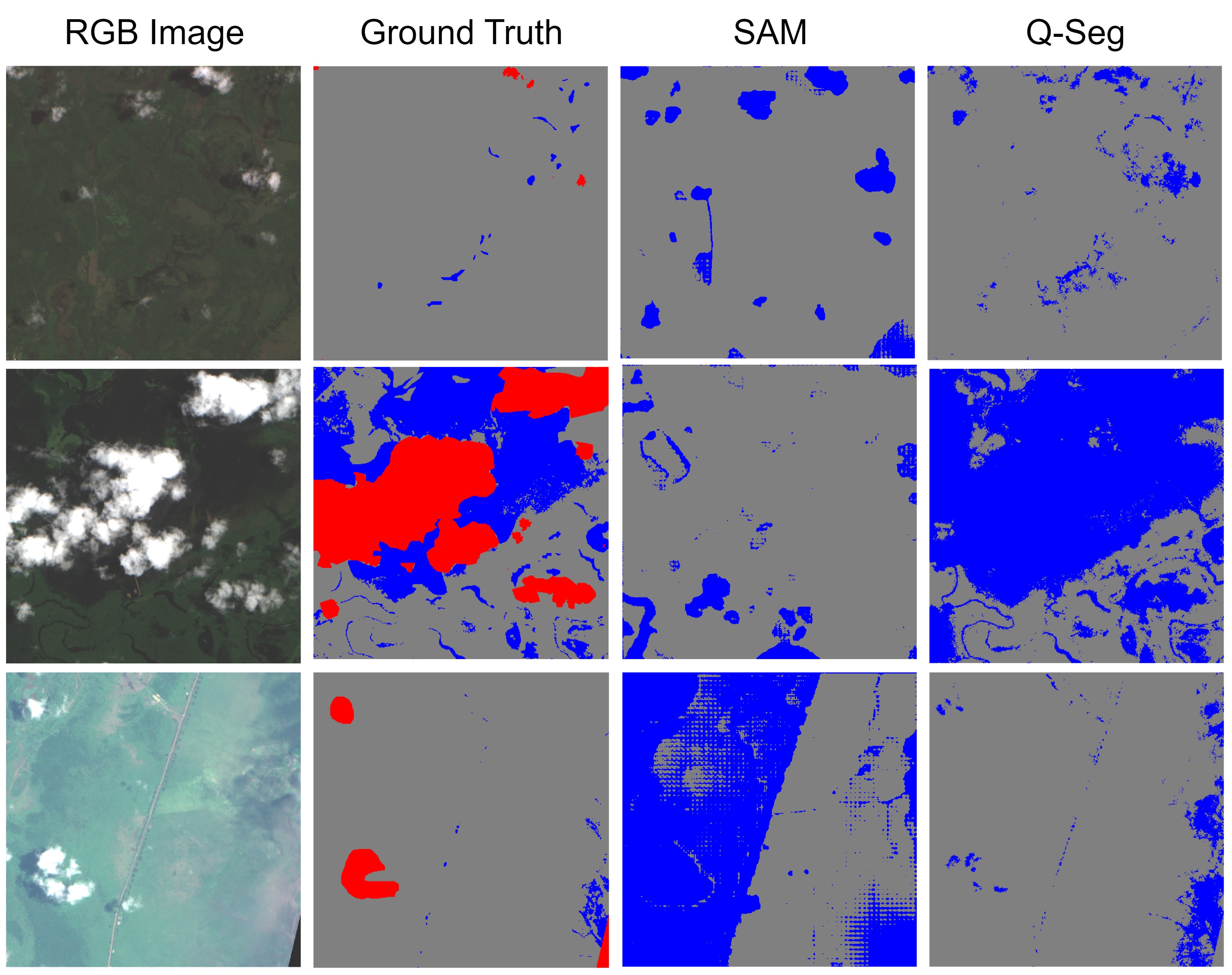}
\caption{Visualization of RGB images, expert-labeled ground truth masks, \textit{Q-Seg}, and \textit{SAM} segmentation results in Flood Mapping. The segmentation differentiates water (blue), dry land (grey), and uncertain areas (red) affected by cloud coverage. \textit{Q-Seg} is observed to be over-segmenting in comparison to the ground truth, which leads to higher precision and shows inconsistent behavior on clouds, suggesting possibilities for improvement in the preprocessing techniques.}
\label{fig: flood mapping results}
\end{figure}

To evaluate the performance of \textit{Q-Seg} in terms of the quality of the segmentation, we compare the results with both \textit{GBDT} and \textit{SAM}. The output masks can be seen in Figure \ref{fig: flood mapping results}, and the quantitative results are reported in Table \ref{tab: flood mapping metrics}.

\begin{table}[ht]
\centering
\begin{tabularx}{0.47\textwidth}{
  >{\hsize=0.232\hsize\linewidth=\hsize\columncolor{white}}X
  >{\hsize=0.200\hsize\linewidth=\hsize\columncolor{rowcolor}}X
  >{\hsize=0.180\hsize\linewidth=\hsize\columncolor{rowcolor}}X
  >{\hsize=0.180\hsize\linewidth=\hsize\columncolor{rowcolor}}X
}
\rowcolor{headercolor}
\textcolor{white}{\textbf{Metrics}} & \textcolor{white}{\textbf{GBDT \cite{iselborn2023importance}}} & \textcolor{white}{\textbf{SAM}} & \textcolor{white}{\textbf{\textit{Q-Seg}}} \\
\midrule
IoU & \textbf{0.888} & 0.330 & 0.775 \\
Accuracy & \textbf{0.913} & 0.463 & 0.828 \\
Recall & \textbf{0.993} & 0.340 & 0.780 \\
Precision & 0.892 & 0.887 & \textbf{0.990} \\
F1-score & \textbf{0.935} & 0.393 & 0.861 \\
\bottomrule
\end{tabularx}
\caption{Performance Comparison of the supervised approach \textit{GBDT} \cite{iselborn2023importance}, \textit{SAM}, and \textit{Q-Seg} in Flood Mapping usecase. The metrics are evaluated ignoring the unsure pixels in the ground truth.}
\label{tab: flood mapping metrics}
\end{table}

Despite the latest and sophisticated capabilities of \textit{SAM}, it demonstrates substandard performance on the unseen dataset, highlighting the robust behavior of \textit{Q-Seg} when applying case-specific metrics.
As expected, the supervised approach outperforms \textit{Q-Seg} in the majority of metrics, except for \textit{Precision} where \textit{Q-Seg} exhibits better performance, likely due to its tendency to over-segment. 
Nevertheless, \textit{Q-Seg} delivers commendable results, closely trailing the supervised approach despite being an unsupervised method and showing more robust behavior than \textit{SAM} in this context. 
Although \textit{Q-Seg} cannot outperform \textit{GBDT}, it could still be a desirable approach when the quality of the output happens to be tolerable for the downstream application.
Moreover, \textit{Q-Seg} circumvents the requirement of reliable hand-labeled data by domain experts and the extensive process of training a supervised learning model, showcasing its potential as a robust tool in challenging segmentation scenarios.

\subsubsection{Scalability}

In this section, we delve into the scalability of our method, particularly in the context of current quantum annealing technologies and their near-term advancements. Our focus is on assessing the adaptability of \textit{Q-Seg} to the evolving landscape of quantum computing, especially in comparison to another quantum annealer-based image segmentation method used for Synthetic Aperture Radar (SAR) images \cite{presles2023synthetic}.
A critical aspect of our analysis is the compatibility of \textit{Q-Seg} with the topological structure of the \textit{D-Wave Advantage} quantum annealers.
Unlike the approach in \cite{presles2023synthetic}, \textit{Q-Seg} demonstrates a better utilization of the quantum hardware capabilities by efficient embedding of binary variables in the QUBO formulation to the physical qubits of the annealer hardware.
As the structure of the graphs used in both approaches are known and \textit{D-Wave} provides the couplers among the physical qubits available in the \textit{Advantage} annealer, we can find the embedding that will be used for solving the QUBO.
Thus, we can deduce the number of physical qubits and eventually reveal the scalability of these approaches.

\begin{figure}[ht]
\centering
\includegraphics[width=0.47\textwidth]{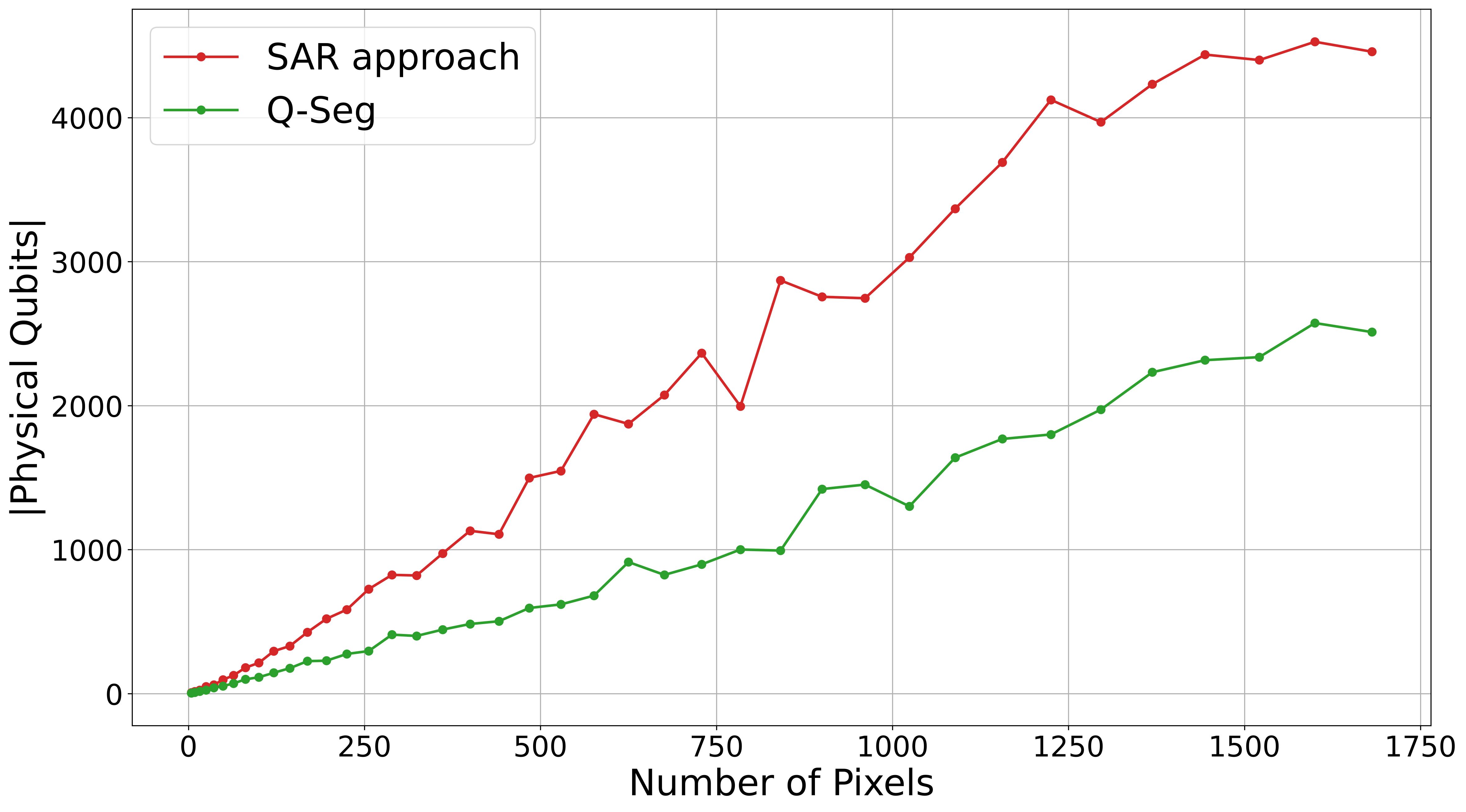}
\caption{Comparison of physical qubit required for \textit{Q-Seg} and SAR segmentation \cite{presles2023synthetic} on the \textit{D-Wave Advantage} with varying image sizes. The qubit requirements are estimated using \textit{minorminer} tool integrated into $dimod$ library, given the logical qubit interactions (quadratic terms in QUBO problem) and the available couplers in the annealer hardware.}
\label{fig: q-seg vs arxiv}
\end{figure}

The results, as shown in Fig. \ref{fig: q-seg vs arxiv}, denote a substantial reduction in the physical qubits required for \textit{Q-Seg} compared to the method in \cite{presles2023synthetic}. 
This efficiency arises from \textit{Q-Seg}'s problem formulation: for an image with $n$ pixels, it necessitates $n$ vertices, each with a maximum degree of $4$.
In contrast, the method in \cite{presles2023synthetic} utilizes $n+2$ vertices, with the terminal vertices having a degree of $n$.
The edges in the graph represent logical qubit interactions, where assigning a logical qubit with extensive interactions to a physical qubit in the QPU poses challenges, requiring multiple physical qubit clones for effective representation.
This directly impacts the annealer performance, as an ideal embedding involves a one-to-one mapping of logical to physical qubits.
Furthermore, the solution space complexity for the method in \cite{presles2023synthetic} is $2^{n+2}$, four times that of \textit{Q-Seg}.
Given that there are only up to $2^n$ possible binary segmentations of the image, all encompassed within the solution space of \textit{Q-Seg}, it suggests a prevalence of redundant solutions in the method of \cite{presles2023synthetic}.

\section{Conclusion} \label{sec: conclusion}

In this paper, we introduced \textit{Q-Seg}, a novel unsupervised segmentation algorithm designed specifically for existing quantum annealers. By efficiently reformulating the segmentation task as a minimum-cut problem in a graph and subsequently as a QUBO problem, our proposed quantum approach efficiently leverages the hardware topology of \textit{D-Wave} annealers, harnessing the advantages of quantum computing. 
Along with enhancing the scalability, it ensures the generation of high-quality solutions using near-term quantum devices.

Our experiments on synthetic data with simulated images of varying sizes demonstrate a significant runtime advantage of \textit{Q-Seg} in processing large-scale images compared to the state-of-the-art optimizer \textit{Gurobi} (Fig. \ref{fig: practical runtimes}). 
Notably, the runtime efficacy is achieved without a significant impact on the quality (Fig. \ref{fig: annealer vs gurobi quality}).
Further, \textit{Q-Seg} was applied to real-world tasks such as Forest Cover and Flood Mapping to test its applicability.
The results suggest a potential for quantum computing to contribute to image segmentation, especially in scenarios with noisy and unreliable annotations.
The results hint at \textit{Q-Seg}'s potential as a practical tool for certain segmentation tasks without extensive training or expert-labeled data, while also acknowledging the need for further exploration and validation.

The future direction for \textit{Q-Seg} is aimed at broadening its application scope beyond earth observation and enhancing segmentation quality.
Upcoming enhancements include extending the algorithm to address multi-cut partitions, which will allow segmentation into multiple semantic categories beyond binary distinctions.
Discussions on such extensions are included in the appendix.
Additionally, there is an opportunity to explore more intricate graph structures that capture a broader spectrum of spatial information, enriching the segmentation process with greater detail and accuracy.
Concurrently, an exploration into the capabilities of universal gate-based quantum computers for segmentation tasks is underway, focusing on the challenges posed by large-scale real-world images and leveraging the diverse technologies of existing hardware.
These initiatives are driven by the goal of unlocking new capabilities in quantum computing to fully exploit its potential for tasks in computer vision.

\section{Appendix}
\newtext{In this appendix, we discuss several possible extensions to the \textit{Q-Seg} algorithm that could enhance its applicability and performance in various segmentation tasks. These formulations are exploratory and intended to provide a foundation for future research rather than final, fully-tested algorithms.}

\subsection*{Extension to Multiple Classes}

For multi-class segmentation involving \( p \) distinct classes, a potential extension of our approach is to create \( p \) individual binary segmentation tasks. Each task would segregate one particular class from all others, effectively converting a multi-class problem into multiple binary segmentation problems. In this proposed method, we define \( p \) different sets of edge weights, \( w_1, w_2, \ldots, w_p \), where each set \( (k = 1, 2, \ldots, p) \) is specifically calibrated to highlight the boundaries of the \( k \)-th class. The objective is to compute \( p \) min-cut problems, each yielding a binary mask that segments the \( k \)-th class from the rest.

The QUBO formulation for each of these binary segmentation tasks can be represented as:

\[
x^*_{k} = \arg \min_{x} \sum_{1 \leq i < j \leq n} x_{v_{i}} (1-x_{v_{j}}) w_k(v_{i},v_{j})
\]

where \( x = x_{v_{1}}, x_{v_{2}}, \ldots, x_{v_{n}} \) and \( x_{v_{i}} \in \{0,1\} \) for the \( k \)-th task. Here, \( x^*_k \) represents the optimal binary segmentation for the \( k \)-th class. The weight function \( w_k(v_{i},v_{j}) \) encodes the edge weights for segmenting the \( k \)-th class, capturing the distinction between the pixels belonging to the class and those that do not.

By repeating this process for each class, we obtain \( p \) binary masks, each isolating one of the \( p \) classes. These masks can then be combined to construct a final multi-class segmentation of the image, demonstrating the potential flexibility and scalability of \textit{Q-Seg}.
\newtext{However, it is important to note that this multi-class extension is currently theoretical and has not been evaluated experimentally. Future work will focus on testing and validating this approach to assess its practical efficacy and performance.}

\subsection*{Learning the Edge Weights}

An alternative approach for defining the edge weight metric—a critical aspect of \textit{Q-Seg} (or any other graph-based segmentation method), particularly in scenarios with sparse training data. By reverse-engineering the annotated segmentation masks available in such limited datasets, we can derive an empirical methodology for estimating the edge weight function. This strategy not only mitigates the issue of insufficient training data but also enhances the adaptability of \textit{Q-Seg} to diverse applications by providing a tailored formulation for edge weight computation.

\textbf{Problem Formulation:}
Let $I$ be an image with pixels indexed by $i$. Let $M$ be the corresponding binary segmentation mask, where $M_i = 1$ if pixel $i$ is in the foreground and $M_i = 0$ if it's in the background. Let $p(i)$ be the set of neighboring pixels of pixel $i$.

\textbf{Feature Extraction:}
For the evaluation of edge weights between each pair of neighboring pixels $(i, j)$, with $j$ being a member of the neighborhood $p(i)$, a multi-dimensional feature vector $x_{ij}$ is meticulously constructed to encapsulate attributes like similarity/dissimilarity, gradient information, texture features, and statistical descriptors or color space transformations. This approach, by integrating a concise yet comprehensive set of information—brightness, contours, surface patterns, and application-specific criteria—enhances the determination of edge weights for graph-based image segmentation, aiming to accurately capture the complexity and diversity of images.

\textbf{Target Weight Assignment ($w^*$):}
Assigning a target weight $w^*_{ij}$ for each pair $(i,j)$:

\begin{equation}
w^*_{ij} = \begin{cases}
-1 & \text{if } M_i \neq M_j \quad (\text{boundary}) \\
+1 & \text{if } M_i = M_j \quad (\text{no boundary})
\end{cases}
\end{equation}

Here, $w^*_{ij} = -1$ indicates a boundary (low weight for edges between different segments), and $w^*_{ij} = +1$ indicates no boundary (high weight for edges within the same segment). Bipolar labeling enhances learning efficiency and accuracy in models, especially neural networks with $\tanh$ activation. It simplifies edge weight interpretation, aiding in precise boundary and ensuring balanced weight distribution for improved boundary detection and region consistency. This method enhances generalization and optimizes the min-cut algorithm by equally valuing boundary presence and absence.

\textbf{Optimization Problem:}
We aim to learn a function $f(\mathbf{x}_{ij}; \theta)$ that maps the feature vector $\mathbf{x}_{ij}$ to an estimated weight $w_{ij}$, approximating the ideal weight $w^*_{ij}$. The function is parameterized by $\theta$. The objective is to find parameters that minimize a loss function $L$ over our labeled dataset, which quantifies the difference between the predicted weights $w_{ij}$ and the target weights $w^*_{ij}$.

Let's say we have a dataset of $N$ images $\{I^1, I^2, \ldots, I^{N}\}$ and their corresponding binary segmentation masks $\{M^1, M^2, \ldots, M^{N}\}$. For each image $I_k$ and its corresponding mask $M_k$, we extract feature vectors $x^k_{ij}$ and target weights $w^{*k}_{ij}$ for each pair of neighboring pixels $(i,j)$. The loss function $L(\theta)$ for the entire dataset is formulated as the average of the individual losses computed for each image:

\begin{equation}
\begin{split}
L(\theta) = &-\frac{1}{N} \sum_{k=1}^{N} \frac{1}{n_k} \sum_{(i, j)} \left[ w_{ij}^{*k} \log f(\mathbf{x}_{ij}^k; \theta) \right. \\
&\left. + (1 - w_{ij}^{*k}) \log (1 - f(\mathbf{x}_{ij}^k; \theta)) \right]
\end{split}
\end{equation}

where $n_k$ is the number of pixel pairs in image $I_k$, $w^{*k}_{ij}$ is the target weight for the pixel pair $(i,j)$ in image $I_k$, determined by the segmentation mask $M_k$ and, $x^k_{ij}$ is the feature vector for the pixel pair $(i,j)$ in image $I^k$.

\textbf{Model Selection:}
The function $f$ can be modeled using various machine learning algorithms. The choice depends on the dataset's complexity and the features.

\textbf{Using Learned Weights in Min-Cut:}
After training, use $f$ to assign weights in our graph for \textit{Q-Seg}:
\begin{equation}
    w_{ij} = f(\mathbf{x}_{ij}; \hat{\theta})
\end{equation}

where $\hat{\theta}$ are the optimized parameters.

\newtext{These potential extensions require further testing and validation to determine their practical efficacy and performance.}

\section*{Code Availability}

All code to generate the data, figures, analyses, as well as, additional technical details on the experiments are publicly available at \href{https://github.com/supreethmv/Q-Seg}{\textcolor{blue}{https://github.com/supreethmv/Q-Seg}}.

\section*{Acknowledgments}

S. Mysore Venkatesh acknowledges support through a scholarship from the University of Kaiserslautern-Landau.
This work has been partially funded by the German Ministry for Education and Research (BMB+F) in the project QAI2-QAICO under grant 13N15586.

%% file: sec/X_suppl.tex
\break

\textbf{Supreeth Mysore Venkatesh} is a researcher at the German Research Center for Artificial Intelligence and pursuing a Ph.D. at the Technical University of Kaiserslautern, Germany. His research focuses on Quantum Algorithms, Machine Learning, and Complexity Theory. Mysore Venkatesh received his Master's in Mathematics and Computer Science from Saarland University, Germany, in 2022. Contact him at \href{mailto:supreeth.mysore@rptu.de}{supreeth.mysore@\{rptu.de}, \href{mailto:supreeth.mysore@dfki.de}{dfki.de\}}  or \href{https://www.supreethmv.com/}{www.supreethmv.com}.
\\
\\

\textbf{Antonio Macaluso} is a Senior Researcher at the German Research Center for Artificial Intelligence (DFKI), Saarbruecken, Germany, 66123 since 2022. His research interests span quantum algorithms in AI, including supervised and reinforcement learning, and multi-agent systems. Macaluso received his Ph.D. in Computer Science and Engineering from the University of Bologna, Italy in 2021. Contact him at \href{mailto:antonio.macaluso@dfki.de}{antonio.macaluso@dfki.de} or \href{https://www.antoniomacaluso.com}{www.antoniomacaluso.com}
\\
\\

\textbf{Marlon Nuske} received his Master's and PhD degrees in Physics from the University of Hamburg in 2015 and 2020, respectively. He is now working at the German Research Center for Artificial Intelligence (DFKI) in Kaiserslautern as a Senior Researcher leading the Earth and Space Applications team since 2021. His research interests lie in machine learning applications in earth observation, data fusion, hybrid modeling techniques, and physics-aware machine learning. Contact him at \href{mailto:marlon.nuske@dfki.de}{marlon.nuske@dfki.de}.
\\
\\

\textbf{Matthias Klusch} is a Principal Researcher and Research Fellow at the German Research Center for Artificial Intelligence (DFKI) in Saarbruecken, Germany. His research interests include hybrid neuro-symbolic AI, semantic technologies, and quantum AI. Klusch received his habilitation in computer science from Saarland university. 
Contact him at \href{https://www.dfki.de/~klusch/}{www.dfki.de/~klusch}.
\\
\\

\textbf{Andreas Dengel} is a Scientific Director at the German Research Center for Artificial Intelligence (DFKI GmbH), Kaiserslautern, Germany. His research interests include pattern recognition, document understanding, and information retrieval. He received his Ph.D. in computer science from the University of Stuttgart. He is a Fellow of the International Association for Pattern Recognition (IAPR). Contact him at \href{mailto:andreas.dengel@dfki.de}{andreas.dengel@dfki.de}

%% file: main.bbl
\begin{thebibliography}{20}
\providecommand{\natexlab}[1]{#1}
\providecommand{\url}[1]{\texttt{#1}}
\expandafter\ifx\csname urlstyle\endcsname\relax
  \providecommand{\doi}[1]{doi: #1}\else
  \providecommand{\doi}{doi: \begingroup \urlstyle{rm}\Url}\fi

\bibitem[Ayodele(2022)]{10.1007/978-3-031-04148-8_11}
Mayowa Ayodele.
\newblock Penalty weights in qubo formulations: Permutation problems.
\newblock In \emph{Evolutionary Computation in Combinatorial Optimization}, pages 159--174, Cham, 2022. Springer International Publishing.

\bibitem[Comaniciu and Meer(1997)]{609410}
D. Comaniciu and P. Meer.
\newblock Robust analysis of feature spaces: color image segmentation.
\newblock In \emph{Proceedings of IEEE Computer Society Conference on Computer Vision and Pattern Recognition}, pages 750--755, 1997.

\bibitem[Danda et~al.(2019)Danda, Challa, Daya~Sagar, and Najman]{8694009}
Sravan Danda, Aditya Challa, B.~S. Daya~Sagar, and Laurent Najman.
\newblock Revisiting the isoperimetric graph partitioning problem.
\newblock \emph{IEEE Access}, 7:\penalty0 50636--50649, 2019.

\bibitem[Glover et~al.(2022)Glover, Kochenberger, Hennig, and Du]{Glover2022}
Fred Glover, Gary Kochenberger, Rick Hennig, and Yu Du.
\newblock Quantum bridge analytics i: a tutorial on formulating and using qubo models.
\newblock \emph{Annals of Operations Research}, 314\penalty0 (1):\penalty0 141--183, 2022.

\bibitem[Gould et~al.(2009)Gould, Gao, and Koller]{NIPS2009_a7aeed74}
Stephen Gould, Tianshi Gao, and Daphne Koller.
\newblock Region-based segmentation and object detection.
\newblock In \emph{Advances in Neural Information Processing Systems}. Curran Associates, Inc., 2009.

\bibitem[Hadlock(1975)]{doi:10.1137/0204019}
F. Hadlock.
\newblock Finding a maximum cut of a planar graph in polynomial time.
\newblock \emph{SIAM Journal on Computing}, 4\penalty0 (3):\penalty0 221--225, 1975.

\bibitem[Iselborn et~al.(2023)Iselborn, Stricker, Miyamoto, Nuske, and Dengel]{iselborn2023importance}
Kevin Iselborn, Marco Stricker, Takashi Miyamoto, Marlon Nuske, and Andreas Dengel.
\newblock On the importance of feature representation for flood mapping using classical machine learning approaches.
\newblock \emph{arXiv preprint arXiv:2303.00691}, 2023.

\bibitem[Kirillov et~al.(2023)Kirillov, Mintun, Ravi, Mao, Rolland, Gustafson, Xiao, Whitehead, Berg, Lo, Doll{\'a}r, and Girshick]{kirillov2023segany}
Alexander Kirillov, Eric Mintun, Nikhila Ravi, Hanzi Mao, Chloe Rolland, Laura Gustafson, Tete Xiao, Spencer Whitehead, Alexander~C. Berg, Wan-Yen Lo, Piotr Doll{\'a}r, and Ross Girshick.
\newblock Segment anything.
\newblock \emph{arXiv:2304.02643}, 2023.

\bibitem[Kotaridis and Lazaridou(2021)]{kotaridis2021remote}
Ioannis Kotaridis and Maria Lazaridou.
\newblock Remote sensing image segmentation advances: A meta-analysis.
\newblock \emph{ISPRS Journal of Photogrammetry and Remote Sensing}, 173:\penalty0 309--322, 2021.

\bibitem[Larasati et~al.(2022)Larasati, Le, and Kim]{9932103}
Harashta~Tatimma Larasati, Thi-Thu-Huong Le, and Howon Kim.
\newblock Trends of quantum computing applications to computer vision.
\newblock In \emph{2022 International Conference on Platform Technology and Service (PlatCon)}, pages 7--12, 2022.

\bibitem[Monien and Sudborough(1988)]{MONIEN1988209}
B. Monien and I.H. Sudborough.
\newblock Min cut is np-complete for edge weighted trees.
\newblock \emph{Theoretical Computer Science}, 58\penalty0 (1):\penalty0 209--229, 1988.

\bibitem[Ng et~al.(2001)Ng, Jordan, and Weiss]{NIPS2001_801272ee}
Andrew Ng, Michael Jordan, and Yair Weiss.
\newblock On spectral clustering: Analysis and an algorithm.
\newblock In \emph{Advances in Neural Information Processing Systems}. MIT Press, 2001.

\bibitem[Presles et~al.(2024)Presles, Enderli, Burel, and Baghious]{presles2023synthetic}
Timothé Presles, Cyrille Enderli, Gilles Burel, and El~Houssaïn Baghious.
\newblock Synthetic aperture radar image segmentation with quantum annealing.
\newblock \emph{IET Radar, Sonar \& Navigation}, n/a\penalty0 (n/a):\penalty0 1–13, 2024.

\bibitem[Ronneberger et~al.(2015)Ronneberger, P.Fischer, and Brox]{RFB15a}
O. Ronneberger, P.Fischer, and T. Brox.
\newblock U-net: Convolutional networks for biomedical image segmentation.
\newblock In \emph{Medical Image Computing and Computer-Assisted Intervention (MICCAI)}, pages 234--241. Springer, 2015.
\newblock (available on arXiv:1505.04597 [cs.CV]).

\bibitem[Solanki and Ingle(2018)]{8933594}
Bhumika Solanki and Maya Ingle.
\newblock Performance evaluation of thresholding techniques on modi script.
\newblock In \emph{2018 International Conference on Advanced Computation and Telecommunication (ICACAT)}, pages 1--6, 2018.

\bibitem[Tse et~al.(2018)Tse, Mountney, Klein, and Severini]{tse2018graph}
Lisa Tse, Peter Mountney, Paul Klein, and Simone Severini.
\newblock Graph cut segmentation methods revisited with a quantum algorithm.
\newblock \emph{arXiv preprint arXiv:1812.03050}, 2018.

\bibitem[Venkatesh et~al.(2022)Venkatesh, Macaluso, and Klusch]{bilp-q}
Supreeth~Mysore Venkatesh, Antonio Macaluso, and Matthias Klusch.
\newblock Bilp-q: Quantum coalition structure generation.
\newblock In \emph{Proceedings of the 19th ACM International Conference on Computing Frontiers}, page 189–192, New York, NY, USA, 2022. Association for Computing Machinery.

\bibitem[Venkatesh et~al.(2023)Venkatesh, Macaluso, and Klusch]{gcs-q}
Supreeth~Mysore Venkatesh, Antonio Macaluso, and Matthias Klusch.
\newblock Gcs-q: Quantum graph coalition structure generation.
\newblock In \emph{Computational Science -- ICCS 2023}, pages 138--152, Cham, 2023. Springer Nature Switzerland.

\bibitem[Wang et~al.(2023)Wang, Cheng, Chen, Shao, Zhu, Wu, Liu, and Zhu]{Wang_2023}
Yan Wang, Jian Cheng, Yixin Chen, Shuai Shao, Lanyun Zhu, Zhenzhou Wu, Tao Liu, and Haogang Zhu.
\newblock {FVP}: Fourier visual prompting for source-free unsupervised domain adaptation of medical image segmentation.
\newblock \emph{{IEEE} Transactions on Medical Imaging}, pages 1--1, 2023.

\bibitem[Yi and Moon(2012)]{yi2012image}
Faliu Yi and Inkyu Moon.
\newblock Image segmentation: A survey of graph-cut methods.
\newblock In \emph{2012 international conference on systems and informatics (ICSAI2012)}, pages 1936--1941. IEEE, 2012.

\end{thebibliography}
